# Cognitive Bias for Universal Algorithmic Intelligence

Alexey Potapov, Sergey Rodionov, Andrew Myasnikov, Galymzhan Begimov

AIDEUS, Russia
{potapov, rodionov, catsith, begimov}@aideus.com

**Abstract.** Existing theoretical universal algorithmic intelligence models are not practically realizable. More pragmatic approach to artificial general intelligence is based on cognitive architectures, which are, however, non-universal in sense that they can construct and use models of the environment only from Turing-incomplete model spaces. We believe that the way to the real AGI consists in bridging the gap between these two approaches. This is possible if one considers cognitive functions as a "cognitive bias" (priors and search heuristics) that should be incorporated into the models of universal algorithmic intelligence without violating their universality. Earlier reported results suiting this approach and its overall feasibility are discussed on the example of perception, planning, knowledge representation, attention, theory of mind, language, and some others.

**Key words:** Universal Agents, Cognitive Architectures, Perception, Attention, Representations, Communication Priors, Value Functions

## 1 Introduction

In Searle's definition, weak AI is distinguished from strong AI (SAI), which can think in exactly the same sense as human beings think. Notion "human-level AI" is also frequently used now. Both these notions are very anthropomorphic and rather vague, because they refer to the poorly apprehensible phenomenon. Clearer distinction is between specialized and general intelligence without references to human intelligence.

In this context it is somewhat surprising that the mainstream approach to the artificial general intelligence (AGI) is based on cognitive architectures [1], most of which try to mimic different aspects of human intelligence starting from the lowest levels in the case of emergent architectures and ending with the high-level cognitive functions in the case of symbolic architectures.

Motivation for cognitive architectures is also clearly explained referring to the human brain: "Different parts of the brain carry out various functions, and no one part is particularly intelligent on its own, but working in concert within the right architecture they result in human-level intelligence… On the other hand, most of the work in the AI field today is far less integrative than what we see in the brain." [2]

Of course, integrative investigations are very useful. But are they aimed at AGI or SAI? Of course, some AI systems can be both strong (human-like) and general. But there are apparent methodological differences between these two approaches.

Possibility of achieving AGI via cognitive architectures is supported by the opinion: "As a result, no one knows what level of intelligence could be achieved by taking an appropriate assemblage of cutting-edge AI algorithms and appropriately integrating them together in a unified framework, in which they can each contribute their respective strengths toward achieving the goals of an overall intelligent system." [2]

However, some doubts can be expressed, because fundamental problems still remain in subareas of AI, and these problems cannot be solved only by integration. It can be seen on the example of machine learning. Existing practical algorithms are still non-universal as being compared with Solomonoff induction [3]. Each algorithm works in Turing-incomplete model space, and cannot extract arbitrary regularity from data. Thus, any system built on the base of such algorithms will be limited in the representable concepts. In some sense, these systems can only do what they are programmed to do.

The following analogy can be used. It was the mathematical model of universal machine with ability to execute arbitrary algorithm or to emulate any other specialized machine that enabled creation of computers. Before this, each automaton implicitly embodied a single particular algorithm and was non-universal in this sense. Universal artificial intelligence should relate to the present cognitive architectures in the same way as computers relate to automata of previous centuries.

Thus, we claim that no combination of non-universal (weak) cognitive functions will result in universal (strong) intelligence. Such combinations will bring yet other (weak) cognitive architectures. It is not bad to refer to the organization of the human brain, but it is simply not enough (for example, the question about "algorithms of the brain" even could not be put before the mathematical theory of algorithms and the universal machine were discovered). This organization should be described in mathematical terms relevant to the universal intelligence.

Models of universal algorithmic artificial (UAI) intelligence do exist [4, 5]. Remarkably, they have almost nothing in common with the cognitive architectures. However, these models are not constructive (they require infinite, or practically infinite computational resources).

Unsurprisingly, these models are criticized for their impracticality, because they don't satisfy the notion of efficient intelligence given in [6] (or of efficient pragmatic general intelligence [7]). Indeed, it is very easy e.g. to play chess possessing infinite computational resources. However, brute force algorithms are not smart in some intuitive sense, although they solve the intellectual task. Thus, it is useful to make distinction between efficient intelligence and mere (inefficient) intelligence. Of course, efficient intelligence is of interest, but the main contribution of the UAI models is their universality.

Of course, the UAI models will be only a small step forward while they remain inefficient. At the same time, most cognitive architectures are as non-universal as the universal models are inefficient. They are situated on the opposite ends of the interval (efficient, non-universal) – (universal, inefficient). Necessity to combine efficiency of cognitive architecture and universality of the UAI models is obvious. However, "the general mathematical theory of AGI, though it has inspired some practical work, has not yet been connected with complex AGI architectures in any nontrivial way" [8]. We believe that this connection is necessary in the first place.

One can either try to introduce universality into the cognitive architectures or to make models of UAI efficient. In this paper, we consider the second way, but the first way is also possible. Cognitive functions are interpreted as metaheuristics and prior information, which make universal intelligence efficient and pragmatic relative to our world.

In several following papers, we give a draft on how one can mathematically introduce such metaheuristics as representations and uncertainty into the models of universal algorithmic intelligence. All other components of cognitive architectures can be and should be incorporated into these models with minimum losses of universality, but with maximum improvements of efficiency. In this paper, we discuss feasibility of formalization of cognitive functions as metaheuristics increasing efficiency of UAI.

## 2     Ideal Minimal Intelligence

The mentioned Solomonoff universal induction can be rather naturally extended to the models of universal algorithmic agents. AIXI is the most well known models thanks to the detailed analysis of its universality and loss bounds [4]. Its time-restricted version is called AIXItl, which obvious drawback is inefficiency (optimality up to some constant slowdown factor [9]).

Some additional doubts can be expressed concerning completeness of this model. Imagine two AIXItl agents playing rock-scissor-paper game many times. The agent with higher computational resources will definitely be winning. However, very simple agent making purely random choices can secure draw in average. Environment is always more complex and has more computational resources, and randomness in action choices can be crucial. This is not modeled in AIXI. In addition, prior and invariable value function can be considered as a decrease of universality.

It was pointed out [9] that constant slowdown can be eliminated by self-optimization, and the Gödel machine with such self-optimization was proposed. Self-optimization is indeed the essential part of the universal intelligent agent. However, the Gödel machine performs self-optimization, only when it can formally prove that this will result in increase of future rewards. But inductive nature of agent's decision-making implies that such proofs can be given only in the very limited number of cases. Probably, a theory of more "soft" self-optimization should be developed.

It can be seen that there are some issues in the existing models of universal agents to be solved. We will refer to some imaginary model with the solved issues as Ideal Minimal Intelligence (IMI) in order to avoid discussion if some concrete model is absolutely satisfactory or not.

The IMI-agent has minimum bias towards some specific environment, where it can be inefficient. But this agent performs self-optimization and given enough experience and time can become efficient pragmatic general intelligence meaning that it will be able to achieve goals in this environment optimally using computational resources. However, it should be pointed out that even if IMI-agents are asymptotically optimal without any multiplicative and additive slowdown factors, it is not enough, because they will require executing too many real-world actions (not just computational operations) in order to acquire information necessary for self-optimization. Probably, the IMI-agent will be required to repeat the whole evolution, which also can be considered as the efficiently self-optimizing search.

Creating real AGI implies that we must essentially accelerate this process. Differences between pragmatic intelligence comprised by many cognitive functions and the simple IMI are entirely conditioned by properties of the real-world environment. This "cognitive bias" contains both prior information for inductive inference and prediction, and heuristics for search procedures. This is a huge bias [10], under which the small IMI core can be almost invisible. However, significance of IMI consists in necessity to introduce this cognitive bias without violating universality. The most direct way to do this is to strictly describe the specific cognitive functions as heuristics extending some IMI model. In the rest part of the paper, we discuss if this approach can be adequately realized.

## 3  Cognitive Bias

It can be argued that the universal intelligence agents can in principle reproduce most forms of human behavior [4] such as supervised learning, board game playing, etc., though the agents are not designed to these specific tasks. Probably, some phenomena such as self-consciousness are not reproduced by these agents and require some additional self-referential structures. However, these issues need separate discussion. Further, we show that cognitive functions don't extend capabilities of IMI-agents, but increase their efficiency in terms of computational resources and learning rates. Several cognitive functions are considered, but this analysis can be extended on the most of them.

*Perception*
IMI models don't include such separate cognitive function as perception. At the same time, natural perception systems are strongly biased towards typical regularities encountered in the real world. This bias is realized in the form of specific information representations and allows for very efficient interpretation of sensory data without exhaustive search.

On the example of perception it is definitely clear that IMI models requiring direct search for algorithmic models of e.g. images with lengths exceeding millions bits (e.g. $\gg 10^{100000}$ alternatives) are absolutely unrealistic. At the same time, it can be noted that natural perceptual systems retain universality. They can learn stimuli with almost arbitrary regularities. And it is always very easy to find stimuli classes, which cannot be learned by machine perception systems relying on the restricted representations. In spite of considerable progress in the fields of robotics, artificial intelligence, machine perception and learning, there is a lack of truly cognitive systems that possess enough generality to deal with unstructured environment [11]. This is why perception systems cannot be implemented as self-contained specialized modules.

The question is how to make the process of model construction (i.e. perception) by IMI more efficient. In a following paper we will describe how one can introduce in IMI such metaheuristic as representation. Representation is the main part of the Representational Minimum Description Length (RMDL) principle, which was developed in attempt to bridge the gap between theoretically ideal universal induction and practical applications of the information-theoretic criteria in computer vision [12]. This principle is derived from the necessity to decompose the task of model construction for the complete sensory history into almost independent subtasks. Because the summed complexity of models constructed in the subtasks is much larger than the complexity of the holistic model constructed for the entire history, direct decomposition is inadmissible. However, if one extracts mutual information from data fragments being independently described and uses this mutual information as prior information for subtasks, summed conditional complexity of models for data fragments can be much closer to the complexity of the holistic model. This mutual information can be treated as a representation.

Hierarchical decompositions are more efficient. Such representations are rather common in machine perception methods. But they are usually introduced heuristically. Moreover, intensive hierarchical decomposition leads to decrease of quality of the models constructed within these representations. Suppression of this drawback can be performed by introducing adaptive resonance that can also be incorporated into information-theoretic induction [13]. Additionally, Solomonoff induction requires practically impossible summation over infinite number of models. Reduction of the algorithmic probability to the finite (typically small) number of models is the unavoidable heuristic. However, one should not simply reduce the number of models, but should introduce uncertain models, which contain simplified information about other possible models. Some mathematical details will be given in another paper.

One can also claim that investigation of particular representations should be guided by criteria derived from the universal induction meaning that researchers also perform optimization in fashion of universal induction. And automatic construction of representations can be considered as an element of self-optimization of IMI.

The RMDL principle is the distinct example of the possible extension of IMI models with the almost unavoidable heuristic of induction task decomposition. We believe that bridging the gap between the other cognitive functions and IMI models in the similar way is the straightest way to AGI.

*Planning*

The notion of representation was formally introduced in the framework of Solomonoff induction to decompose the model construction process. However, the task of sequential decision making also has extremely large computational complexity. The human brain uses some cognitive functions, which can be considered as search metaheuristics. The most obvious one is planning.

Brute force algorithms don't use planning. Interestingly, modern chess-playing programs also don't use planning, while human players almost obligatory rely on a plan while developing their game [14].

One of the origins of planning consists in the possibility to reuse results of search performed on the previous steps. Indeed, plans are built in advance, and then they are only adjusted implying that the search tree is not constructed from scratch in each moment of time. This strategy can be easily incorporated into IMI models, but natural planning is much more sophisticated.

Humans construct plans and perform search in terms of some generalized actions. More distant plans are, more abstract actions are involved. Usage of generalized actions is an obvious heuristic. These actions also form a kind of representation, but it cannot be directly put in the universal induction framework. In practice, such representations are specified a priori, and particular planning algorithms are developed for them. This is insufficient for AGI.

On the one hand, search and optimization research areas including heuristic programming, simulated annealing, genetic algorithms, and other fields are rather developed in the classical AI. On the other hand, there is still no general solution of the search problem. Most likely, there cannot be such solution except some universal self-optimization, because different heuristics and specific search methods suit better for different tasks.

At present, there is no theory of efficient pragmatic general self-optimization capable of invention of arbitrary search heuristics. However, this search (even being given) should also be speeded up using some very general metaheuristics. Otherwise it will not be pragmatic.

*Knowledge*

Knowledge plays the prominent role in human intelligence. At the same time, there is no knowledge representation in IMI, which constructs holistic models of its history without explicitly extracting knowledge from them. Actually, knowledge is sometimes considered simply as the highest level in the hierarchical models of perception and control (e.g. knowledge level of a vision system). In this context, not much can be added to the ideas discussed in the sections devoted to perception and planning. However, knowledge representations are also conditioned by social interactions, which will be discussed later.

In general, knowledge representations can probably be discovered during self-optimization of IMI, but this process requires extremely long-term interactions. Useful representations of the world abstracted from specific modalities can additionally bootstrap transformation of IMI into pragmatic efficient AGI. But, again, these representations shouldn't restrict universality of IMI.

*Memory*

Memory is one of the main components of the most cognitive architectures. At the same time, IMI models start with storing raw history while human memory has very elaborated structure. It was already pointed out [7] that memory (as the cognitive function) is absent in the models of universal intelligence due to supposed unlimited resources, but memory is the essential element of developing these models into efficient pragmatic general intelligence.

Apparently, it is too wasteful to build models of environment at each time moment. These models can be stored and retrieved, when necessary. Structure of memory partially originates from structure of models. For example, semantic and episodic memories probably correspond to stored representations and models of the specific data described within these representations. Chunks and transfer learning can also be explained in terms of induction task decomposition.

Memory is tightly connected with perception, representations, planning and so on. But it also adds the aspect of incremental learning. Both retrieval and adjustment of models stored in memory should be formally introduced into IMI models to increase their efficiency without violating universality. And, of course, implementations of memory in cognitive architectures cannot be universal, if they rely on non-universal induction.

*Attention*

Attention is the diverse phenomenon. However, its origin in management of limited resources is evident. For example, visual attention is aimed at the most informative or significant (in terms of value functions) part of a scene meaning that this part is thoroughly analyzed using more resources than those allocated for other parts. Of course, allocation of resources while solving other cognitive tasks also can be considered as attention.

There are different models of attention for cognitive architecture (e.g. [15, 16]). Even simple universal solvers (e.g. [17]), which take computational complexity into account, try to optimally allocate resources among different hypothesis under consideration. Apparently, more elaborated attention mechanisms should be presented in the efficient pragmatic AGI. However, details of these mechanisms greatly depend on other parts of a cognitive architecture. Thus, the substantive model of attention should be developed jointly with the resource-limited extension of IMI models.

*Motivation*

One of the difficult questions for the theories of universal intelligence is the question about motivation. As it was mentioned, assumption of the invariable value function given a priori is restrictive. Imagine that the agent has a "simple" goal – to survive. What value function should be used? The direct "survival" value function will have only two values corresponding to the states "alive" and "dead". Such the value function can be optimized only by evolution of many agents, because single agent without some specific prior knowledge is unable to predict the "dead" state, which has not been ever encountered.

More complex value functions are survival heuristics. Pain tells us that we are coming nearer to the "dead" state. Pleasure tells us that we are moving away from it. But these prior heuristics are not intelligent and lead to mistakes, when we need to overcome pain or to avoid pleasure to stay alive, and this only can be done thanks to the abstract fear of death. It is very difficult to specify a good value function prior to intelligence.

Not only pain and pleasure, but also different emotions can be considered as additional survival heuristics. They help intelligence to learn good survival strategies quicker. For example, the intelligent agent without particularly design curiosity may learn knowledge-seeking strategy, if it helps to survive. But this agent at least should remain alive until this strategy is learned. Apparently, curiosity factor can be included a priori into the value function as the survival heuristic. And the theory of how to formally describe curiosity or creativity or aesthetic reward already exists [18].

But even if we include some survival heuristics, the agent will still be motivated not to survive, but to maximize the given value function. In contrast, the natural intelligence is not the system of maximization of the given value function, but it is the one large survival heuristic. It is probably not so bad to be curious by itself [19], but the problem is to specify such value functions that the agent's behavior will correspond to our expectations. This problem is discussed in more details on the example of safe or friendly AGI [20]. Indeed, one cannot precisely specify friendliness (as well as survival) a priori as a value function external to the unbiased intelligence. Thus, the AGI should be very biased a priori or should learn better value functions during life-time. However, investigations of the problem of learning the value functions by universal algorithmic agents have started only recently (e.g. [21]).

*Social interactions*

Interactions with other intelligent agents constitute very significant part of the environment. These agents are very complex, so the inductive reconstruction of the appropriate models of other agents will require very long-term interactions in the real world and vast computational resources. Apparently, some theory of mind should be incorporated into efficient pragmatic general intelligence. But it should be added as an element of the bias that shifts universal priors, but doesn't restrict the model space.

Social interactions are not reduced only to predicting behavior (or reconstructing models) of other agents as the part of environment. Of course, social agents interact also via the interface of sensory inputs and actuating outputs, but they can communicate fragments of models of environment, behavior policies, and even value functions. It is actually society that accumulates complex value functions, inductive bias and search heuristics (in the form of ethics, science, etc.) thanks to sharing information and computational resources between agents. Unbiased universal agents can learn to acquire this information from society given enough time (however, it may not be true for value functions). But efficient pragmatic general intelligence should have this ability a priori meaning that it is biased towards social environments [22] or has communication prior [23].

Essential (but not the only) aspect of social interactions is language. Few issues about language in the context of universal agents were considered, including importance of the two-part coding [22] that allows agents to efficiently communicate regular parts of models separated from noise. However, many other issues (such as symbol grounding) are still waiting to be thoroughly considered in this context.

One additional significant aspect of multi-agent interactions is that the environment appears to be much more complex and computationally powerful than the agent. This aspect is not a heuristic or prior, but it should be considered in the basic IMI models.

## 4     Conclusions

Cognitive functions can be properly designed as an inductive bias and search heuristics necessary for the universal algorithmic intelligence to become efficient pragmatic general intelligence (acting in the specific real-world environment using limited resources and relatively short learning time).

Some cognitive functions have already been suchwise represented separately in different papers. Mathematical descriptions extending the models of the universal algorithmic intelligence have also been given for some of them. We believe that the way to AGI lies through systematic investigations in this direction, which are still almost absent. In this paper, a step towards it is taken.

All cognitive functions should be considered and described jointly, because they are interconnected. However, these functions consist of separable metaheuristics, which mathematical models can be introduced consequently. All heuristics and priors depend on the environment properties, but some of them are tightly connected to the task being solved by the intelligent agent. Decomposition of the prediction, decision making, and self-optimization tasks relies on such primary metaheuristics, which define general structure of memory and knowledge representations. Finer details of this structure greatly depend on the bias conditioned by the real-world environment. In particular, specific sensorimotor representations and social priors constitute this bias.

Recent results of different authors on formalization of cognitive functions within the framework of the universal algorithmic intelligence encourage, and the program of systematically introducing the grounded cognitive bias seems to be feasible.